\documentclass[10pt,twocolumn,letterpaper]{article}

\usepackage{iccv}
\usepackage{times}
\usepackage{epsfig}
\usepackage{graphicx}
\usepackage{amsmath}
\usepackage{amssymb}

\usepackage{graphicx}
\usepackage{amsmath}
\usepackage{amssymb}
\usepackage{booktabs}
\usepackage{bm}
\usepackage{makecell}

\usepackage[utf8]{inputenc} 
\usepackage[T1]{fontenc}    
\usepackage{url}            
\usepackage{booktabs}       
\usepackage{amsfonts}       
\usepackage{nicefrac}       
\usepackage{microtype}      
\usepackage{xcolor}         
\usepackage{pifont}
\usepackage{caption}
\usepackage{multirow}
\usepackage{subcaption}
\usepackage{float}
\usepackage[ruled, linesnumbered]{algorithm2e}
\usepackage{graphicx,calc}
\usepackage{multicol}

\usepackage{amsmath}

\usepackage[pagebackref=true,breaklinks=true,letterpaper=true,colorlinks,bookmarks=false]{hyperref}

\iccvfinalcopy 

\def\iccvPaperID{3618} 

\ificcvfinal\fi

\begin{document}

\title{DIR-AS: Decoupling Individual Identification and Temporal Reasoning for Action Segmentation}

\author{Peiyao Wang, Haibin Ling\\
Department of Computer Science,  Stony Brook University\\
{\tt\small \{peiyaowang, hling\}@cs.stonybrook.edu}
}
\maketitle
\ificcvfinal\thispagestyle{empty}\fi

\begin{abstract}
Fully supervised action segmentation works on frame-wise action recognition with dense annotations and often suffers from the over-segmentation issue. Existing works have proposed a variety of solutions such as boundary-aware networks, multi-stage refinement, and temporal smoothness losses. However, most of them take advantage of frame-wise supervision, which cannot effectively tackle the evaluation metrics with different granularities. 
In this paper, for the desirable large receptive field, we first develop a novel local-global attention mechanism with temporal pyramid dilation and temporal pyramid pooling for efficient multi-scale attention. Then we decouple two inherent goals in action segmentation, \ie, (1) individual identification solved by frame-wise supervision, and (2) temporal reasoning tackled by action set prediction. Afterward, an action alignment module fuses these different granularity predictions, leading to more accurate and smoother action segmentation. We achieve state-of-the-art accuracy, \eg, 82.8\% (+2.6\%) on GTEA and 74.7\% (+1.2\%) on Breakfast, which demonstrates the effectiveness of our proposed method, accompanied by extensive ablation studies. The code will be made available later. 
\end{abstract}

\section{Introduction}
\label{sec:intro}

\setlength{\abovecaptionskip}{0.1cm}
\begin{figure}[tbp]
  \centering
  \includegraphics[width=0.98\linewidth]{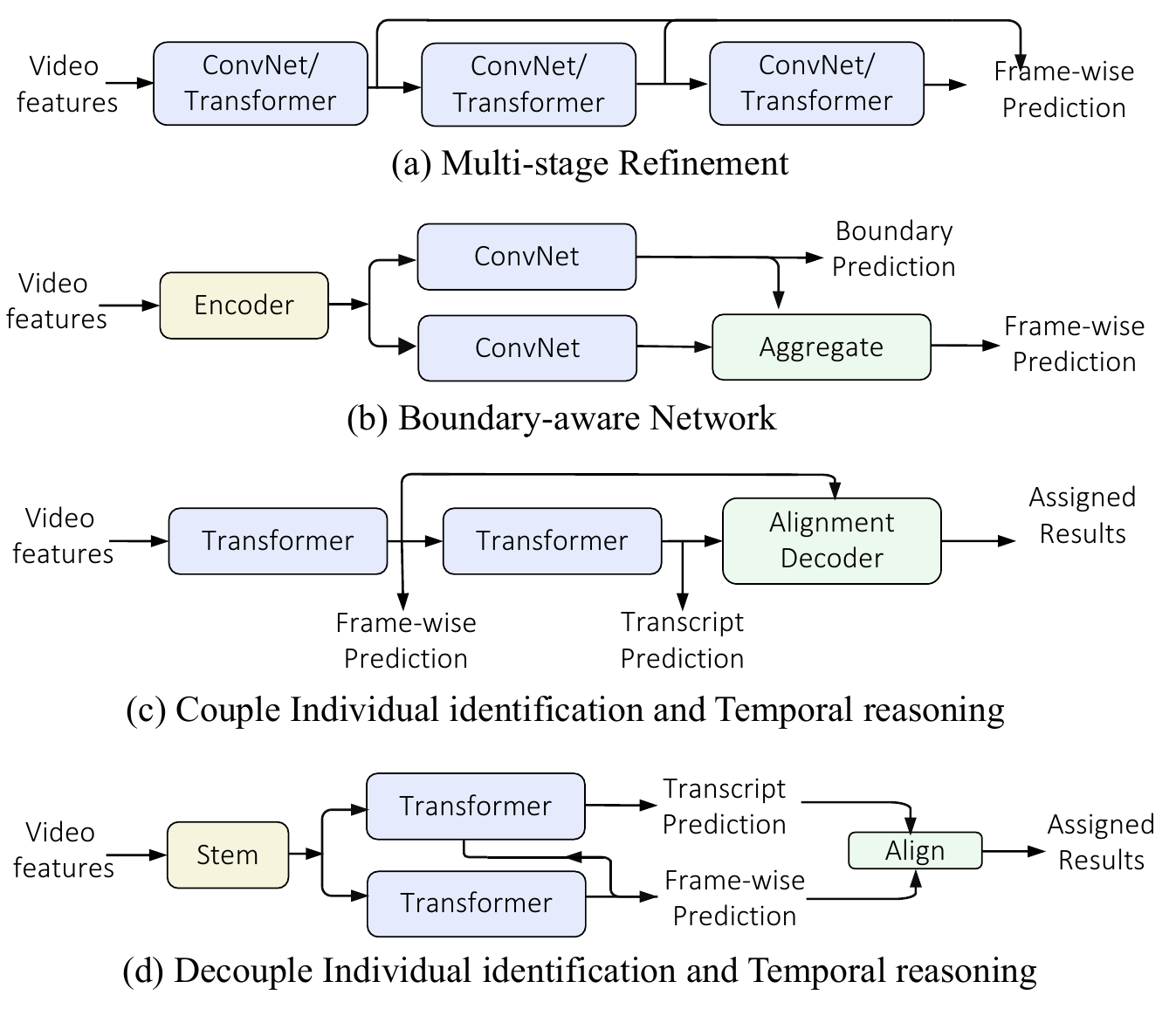}
  \caption{Comparison of existing action segmentation pipelines (a--c) with our proposed decoupling strategy (d). }
  \label{fig:related}
\end{figure}

Temporal action segmentation~\cite{lea2017temporal,wang2020boundary, li2020ms,lei2018temporal} aims to assign each frame an action label in untrimmed videos. It attracts great research interest due to its wide range of applications such as instructional video understanding~\cite{tang2019coin,miech2019howto100m}, human activities analysis~\cite{vallacher1987people}, video surveillance~\cite{collins2000introduction,collins2000system}, and so on. Different from other video understanding tasks working on trimmed videos, such as action recognition identifying an action for a video, dense predictions of action segmentation usually trained by frame-wise supervision easily cause over-segmentation. One potential reason is that a slight variance of consecutive frames between actions results in boundary ambiguity~\cite{wang2020boundary}. Empirically, the networks often suffer from non-convergence on such uncertain boundaries, making the task more challenging than its counterpart semantic segmentation~\cite{strudel2021segmenter,wang2021pyramid,zheng2021rethinking,xie2021segformer} with clear visual boundaries.

To ensure prediction fidelity and continuity, existing benchmarks for action segmentation ~\cite{fathi2011learning, stein2013combining, kuehne2014language} have employed three evaluation metrics: frame-wise accuracy, segment-wise F1 score, and segment-wise Edit score. However, most of previous works~\cite{lea2017temporal, lei2018temporal,yi2021asformer, farha2019ms, wang2020boundary,ishikawa2021alleviating} adopt a single model with frame-wise supervision, in spite of that boundary-aware network~\cite{wang2020boundary,ishikawa2021alleviating}, multi-stage refinement~\cite{farha2019ms, yi2021asformer}, \etc, are introduced to alleviate over-segmentation, demonstrated in Figure~\ref{fig:related}. Instead, we claim that two inherent goals, \ie, individual identification and temporal reasoning, should be decoupled as the former focuses on discreteness while the latter works for continuity. 


How to design an effective network for both individual identification and temporal reasoning then becomes our key motivation. First, we need large Receptive Fields (RFs) since their necessity has been verified in previous works~\cite{lea2017temporal, lei2018temporal, farha2019ms, yi2021asformer}. Then, these two goals should be united into one model rather than separated. Last, boundary ambiguity should be addressed to facilitate training. To this end, we introduce a novel Transformer for efficient and effective action segmentation. 

We first utilize several dilated convolution layers before the Transformer for unifying these two goals in one model. For \textbf{individual identification}, we develop a two-branch attention mechanism with both locality and globalism, where locality is designed for data efficiency and globalism is held for global context modeling. Specifically, the local branch adopts sliding window attention with Temporal Pyramid Dilation (TPD), while the global one leverages Temporal Pyramid Pooling (TPP) on key and value tokens for efficient multi-scale attention. For \textbf{temporal reasoning}, coarse transcripts are obtained from the individual identification module. Then Transformer decoder-based structure takes them as queries to attend to the early features from individual identification for decoupled training. Compared with the frame-wise supervision used in individual identification, we impose action-set supervision on temporal reasoning for accurate transcript output. Then, parameter-free action alignment is used to align these two different granularity outputs. To alleviate boundary ambiguity, we advocate a categorical-level and temporal-level smoothing strategy.

We evaluate our proposed method on two action segmentation benchmarks, \eg, GTEA~\cite{fathi2011learning} and Breakfast~\cite{kuehne2014language}. We achieve state-of-the-art results without multiple stages used in most previous work~\cite{li2020ms, farha2019ms, yi2021asformer}. 

The technical contributions in this paper are summarized as follows:
\textbf{i}) we decouple two critical goals, \ie, individual identification and temporal reasoning for action segmentation, leading to an easier optimization for different granularity evaluation metrics;
\textbf{ii}) we introduce a novel two-branch attention mechanism for efficient global attention on the long-form action segmentation task;
and \textbf{iii}) state-of-the-art performances of multiple evaluation metrics on two benchmarks are achieved by our unified single-stage network.

\section{Related Work}
\subsection{Action segmentation} 
Traditional action segmentation methods use sliding windows with non-maximum suppression~\cite{rohrbach2012database,karaman2014fast}. Other methods~\cite{kuehne2016end,richard2017weakly,pirsiavash2014parsing,donahue2015long,yue2015beyond,singh2016multi,yeung2018every} use RNNs or Conditional Random Fields (CRFs) to model temporal action sequences. ED-TCN~\cite{lea2017temporal} uses a hierarchy of temporal convolutions with dilation to capture long-range temporal patterns with a pooling-upsampling structure but is harmful to fine-grained temporal information. Further, TDRN~\cite{lei2018temporal} uses two parallel temporal streams with deformable temporal convolutions to facilitate local, fine-scale action segmentation and multi-scale context. To alleviate the over-segmentation problem, MS-TCN~\cite{farha2019ms} introduces multistage architecture to capture long-range dependencies. To alleviate the boundary ambiguity problem, many methods~\cite{wang2020boundary,ishikawa2021alleviating} investigate to detect action boundaries to refine results. For example, BCN~\cite{wang2020boundary} leverages cascade strategy to have adaptive RFs for better boundary detection, while ASRF~\cite{ishikawa2021alleviating} proposes action boundary regression to refine ambiguous frames. In addition, graph-based network is employed in~\cite{huang2020improving} for temporal reasoning. 
Recently, ASFormer~\cite{yi2021asformer} introduces Transformer~\cite{vaswani2017attention} with Multi-Head Self-Attention (MHSA) and Multi-Head Cross-Attention (MHCA) to action segmentation, leading to performance breakthrough. Note that a relevant work~\cite{behrmann2022unified} also introduces action set prediction, but they conduct frame-wise and segment-wise predictions at the late stage. However, our motivation in this paper is to decouple these two important tasks by designing an asymmetric architecture at the early stage.

\begin{figure*}[t]
  \centering
  \includegraphics[width=1.0\textwidth]{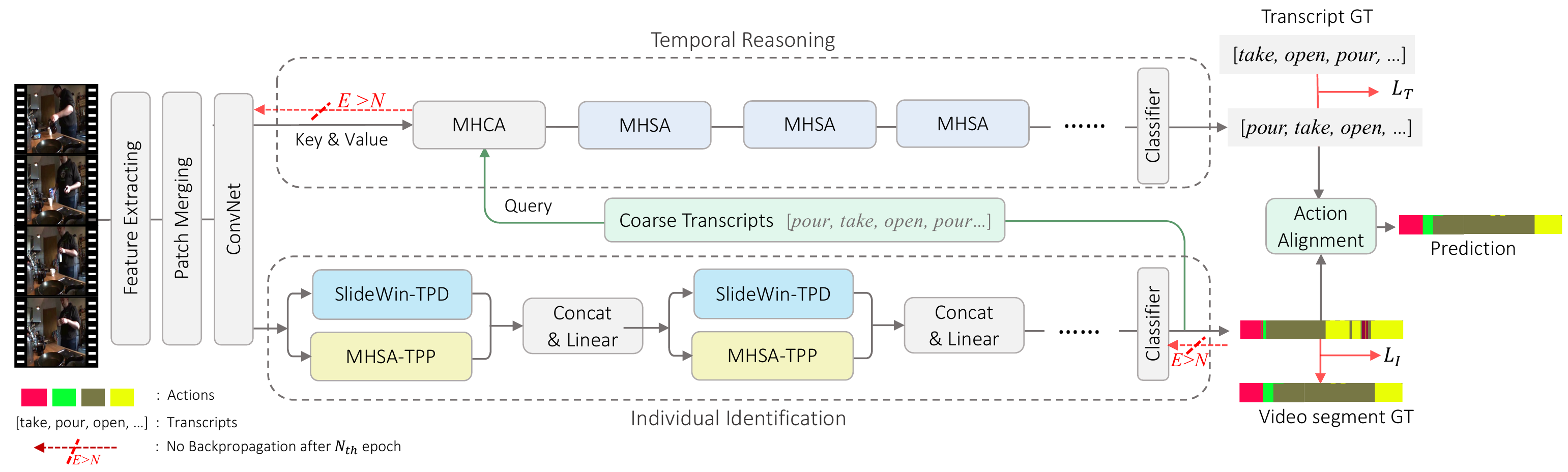}
  \caption{The overall architecture of DIR-AS. Individual identification and temporal reasoning are decoupled at the early stage, \ie, shared ConvNet. In the Individual Identification module, the local branch adopts sliding window attention with Temporal Pyramid Dilation (SlideWin-TPD), while the global branch introduces Multi-Head Self Attention with Temporal Pyramid Pooling (MHSA-TPP). In the temporal reasoning module, multi-head attentions are concatenated sequentially, \ie, an MHCA (Multi-Head Cross Attention) followed by MHSAs. }
  \label{fig:overall}
\end{figure*}

\subsection{Vision Transformer} 
Inspired by the great achievement of transformer~\cite{vaswani2017attention} in natural language processing (NLP), ViT~\cite{dosovitskiyimage} shows that pure-transformer networks can also work well in the vision domain. ViT has given insight into a large amount of follow-up work in the vision domain, such as image classification~\cite{chen2021crossvit,touvron2021training,wang2021pyramid,yuan2021tokens,han2021transformer}, object detection~\cite{carion2020end,zhudeformable,sun2021rethinking}, semantic segmentation~\cite{strudel2021segmenter,wang2021pyramid,zheng2021rethinking,xie2021segformer}. Further, a series of works~\cite{arnab2021vivit,bulat2021space,patrick2021keeping,zha2021shifted,fan2021multiscale,liu2022video} tend to investigate the application of transformer for spatial-temporal learning to capture long-term dependencies. ViViT~\cite{arnab2021vivit} shows four factorized designs along spatial and temporal dimensions. MViT~\cite{fan2021multiscale}, designed for image and video recognition, uses a hierarchical structure to capture a multi-scale pyramid of features and pooling attention to reduce space-time resolution and attention computation. Video Swin transformer~\cite{liu2022video} advocates an inductive bias of locality via window shift. Uniformer~\cite{li2022uniformer} learns local and global token affinity to integrate the merits of both locality in 3D convolutions and long-range dependency in the transformer.


\subsection{Efficient Attention} 
We now review the relevant, efficient attention strategies. First, sliding windows~\cite{parmar2018image, child2019generating, ho2019axial, liu2021swin} are one of the obvious solutions for computation complexity reduction. For instance, Swin-Transformer~\cite{liu2021swin} replaces spatially flattened attention~\cite{dosovitskiyimage} with window-based attention to reduce the quadratic complexity dramatically. Second, low rank based methods~\cite{wang2020linformer, xiong2021nystromformer, tay2021synthesizer} provide another solution. For example, Linformer~\cite{wang2020linformer} dramatically reduces the dimensions of keys and values, leading to a significant computation reduction since it is proportional to the dimensions of both key tokens and query tokens. Third, how to reuse computed features (\eg, memory~\cite{raecompressive, lee2019set}) becomes a powerful alternative. For example, Set Transformer~\cite{lee2019set} uses extra memories to store and reuse the attended features for distant tokens. Last, the kernel-based approximation has also attracted research attention~\cite{choromanskirethinking, katharopoulos2020transformers, peng2021random}.

\section{Approach}

\subsection{Method Overview}

The framework of DIR-AS is illustrated in Figure~\ref{fig:overall}. As previous works~\cite{farha2019ms, yi2021asformer, behrmann2022unified} do, we first extract the features $\bm{L} = \{\bm{l}_t\}_{t=1}^T$ using I3D~\cite{carreira2017quo}, where $\bm {l}_t\in \mathbb{R}^d$ is the $t^{th}$ frame feature and $d$ is the feature dimension. Considering high duplication of video frames, we follow image semantic segmentation~\cite{xie2021segformer} to squeeze the long feature sequence $\bm L$ at the beginning via a convolution layer with down-sampling, yielding a shorter feature sequence $\bm{S} = \{\bm{s}_t\}_{t=1}^{T'}$, where $T'$ is the length of features after down-sampling. Upon the squeezed sequence $\bm S$, we apply several convolution layers with small dilation to yield embedded features $\bm E = \{\bm{e}_t\}_{t=1}^{T'}$ for two reasons: i) CNNs are good at introducing inductive bias, \ie, locality because of shared weights on sliding windows, which is crucial for action segmentation suffering from data insufficiency; and ii) this shared module plays an important role in connecting the individual identification and temporal reasoning rather than using two separated networks.

Two network branches are then designed upon the embedded features $\bm E$ to learn individual identification and transcript prediction features, respectively: i) an efficient local-global network is proposed to learn features with local and global information for better frame-wise classification; ii) another branch projects the embedded features $\bm E$ to an ordered action transcript by cross-attention. Finally, the parameter-free Viterbi algorithm~\cite{kuehne2018hybrid} is adopted to align the predicted transcript to the frame-wise classification for better performance. 



\subsection{Individual Identification}
\begin{figure}[t]
  \centering
   \includegraphics[width=.95\linewidth]{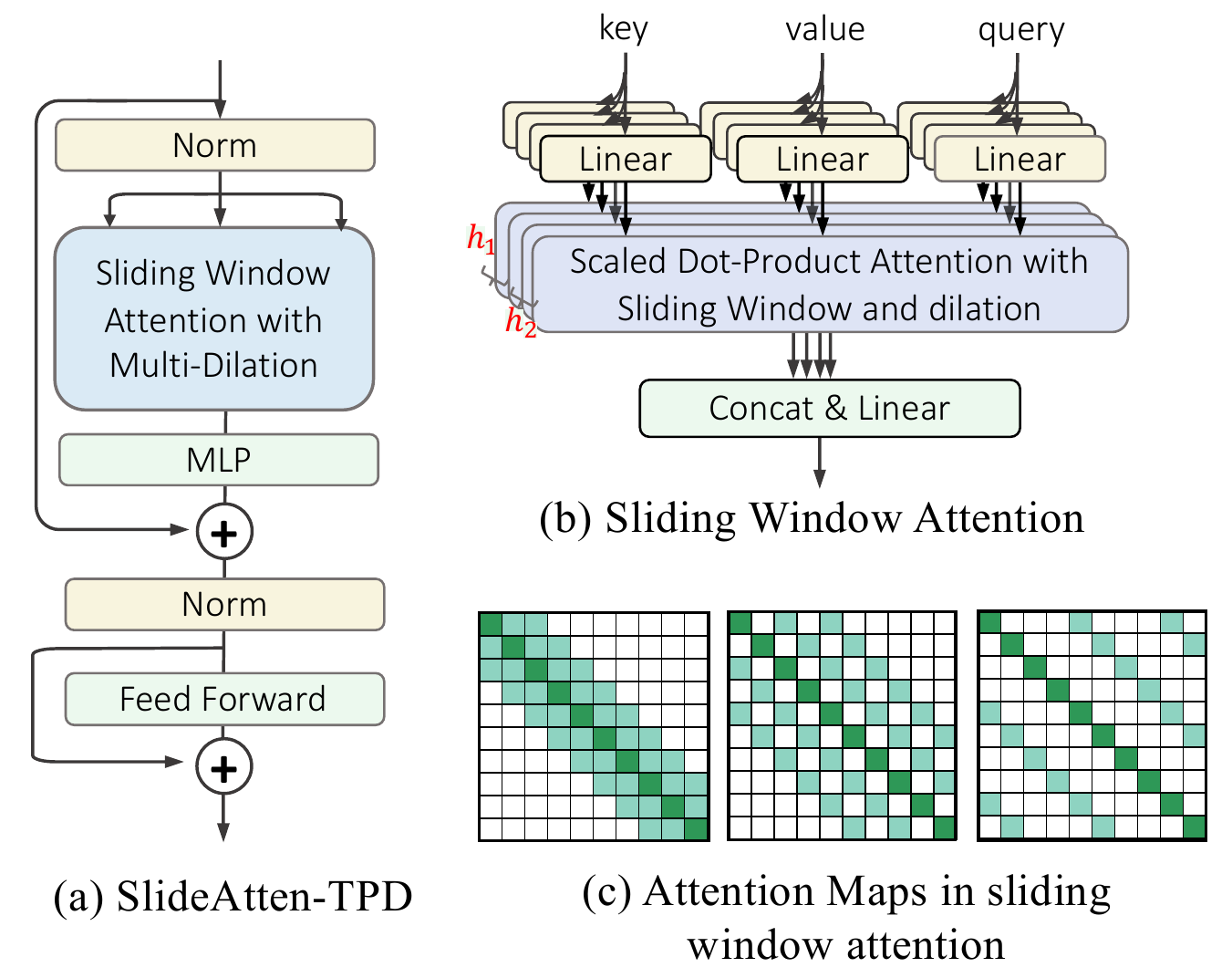}
   \caption{Structure of the local branch. In (c), the attention maps project on the full resolution for to visualize the expanded receptive field. The actual computation cost is calculated only on the green cells. Thus, there are no extra costs for dilation, compared to the vanilla attention (left).}
   \label{fig:local}
\end{figure}

\noindent\textbf{Local branch with temporal pyramid dilation.} Because of Multi-Head Self-Attention (MHSA), Transformer captures a global view over the whole sequence but suffers from square complexity and the lack of inductive bias, \ie, locality~\cite{dosovitskiyimage}. So this vanilla Transformer is not appropriate for extra-long sequence data in action segmentation without sufficiently large training samples. Instead, we use the sliding window attention to introduce locality, with dilation to enlarge the limited RFs caused by the certain window partition. In particular, we extend such an operation into a multi-scale dilation variant realized by multi-heads, which we name Temporal Pyramid Dilation (TPD) and demonstrate the attention in Figure~\ref{fig:local}. MHSA with TPD (MHSA-TPD) has two obvious advantages for action segmentation. First, it captures long and short ranges of contexts while action length varies due to the slow and fast motions. Second, it is efficient since we implement multiple dilation operations on MHSA, resulting in no extra computation costs as common dilated convolution does.

\begin{figure}[t]
  \centering
   \includegraphics[width=0.92\linewidth]{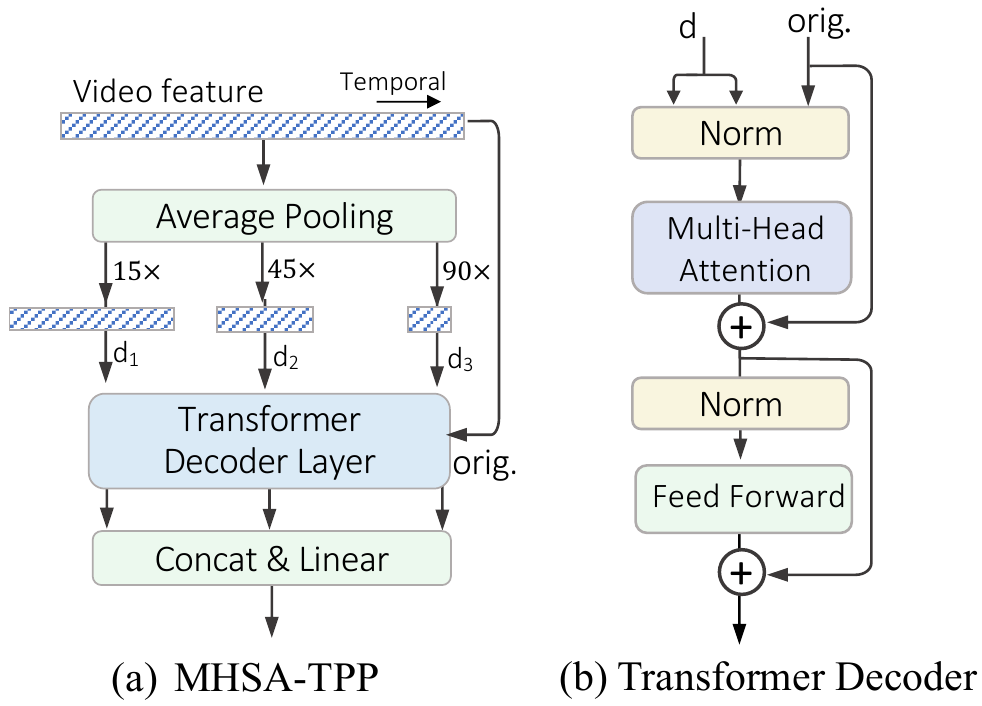}

   \caption{Structure of the global branch. In (a), a Transformer Decoder layer utilizes video features with full resolution as query and downsampled features as key and value, allowing for efficient processing of long-term video data.}
   \label{fig:global}
\end{figure}

\noindent\textbf{Global branch with temporal pyramid pooling.} In addition to the gradually accumulated RFs by the local branch, we provide another global branch to attend to the whole sequence quickly. Specifically, we apply a large $d_s$-fold down-sampling on key and value tokens so that the square complexity $O(T'^2)$ can be dramatically reduced to $O(\frac{T'^2}{d_s})$. Similar to the multi-scale dilation in the local branch, we advocate Temporal Pyramid Pooling (TPP) on the key and value tokens, so it also can capture multi-scale global contexts. As shown in Figure~\ref{fig:global}, we concatenate these attended features on the channel, and it can be seamlessly embedded into a regular Transformer encoder.

As noted in Figure~\ref{fig:overall}, by concatenating the local and global branches on a channel and repeating this module multiple times, we achieve a meaningful representation of slow and fast actions. It is worth noting that we retain the original temporal resolution for all the features in the local-global branches as we intend to keep all the temporal information. Before performing frame-wise classification, we interpolate the down-sampled features to the original resolution by Nearest Neighbor. Other interpolation modes are applicable, and we empirically find that Nearest Neighbor brings better performance as consecutive video frames always hold the same action label.

\subsection{Temporal reasoning}

To convert dense frame-wise features to short action transcript predictions, we leverage a cross-attention where the limited number of query tokens attend the whole sequence, resulting in a short sequence output. Here we consider two options for the query tokens: (1) randomly initialized tokens, and (2) coarse action transcripts accumulated by frame-wise predictions from individual identification. We experimentally find that the latter option brings more performance gain and even improves the frame-wise accuracy thanks to the early shared convolutions. One potential reason is that the conditional cross-attention more dynamically attends the whole sequence sample-by-sample compared to the constant cross-attention for all the samples. More discussions can be found in experiments.

After that, a series of self-attention layers sequentially exchange temporal information for the final action transcript predictions. Since the number of actions varies in each video, we add an extra `End' token to recognize the end of the action transcript. 

\setlength{\abovecaptionskip}{0.1cm}

\noindent\textbf{Action alignment.} In previous works~\cite{li2020ms,yi2021asformer}, the frame-wise results from the individual identification can be found in the final prediction for all the evaluation metrics. However, as aforementioned, one single model may not effectively tackle these evaluation metrics with different granularities. Instead, our decoupled modules can be optimally trained respectively, as illustrated in the following section. Thus, we align the best transcript to the best frame-wise classification for better performance. Specifically, we adopt the Viterbi algorithm~\cite{kuehne2018hybrid} as in~\cite{behrmann2022unified}. It takes the frame-wise probabilities and transcript as inputs and efficiently enumerates all the possible durations for all the action labels in the transcript. The best solution is selected by the highest accumulated probabilities. Then, we can easily convert the assigned durations to frame-wise action predictions as final results. Note that the Edit score of the frame-wise predictions from the alignment is still the same as the best transcript from the temporal reasoning network, as Viterbi only assigns durations without altering action order in the transcript.


\begin{algorithm}[t]
\caption{ Decoupled Training for DIR-AS}
\label{alg:alg1}
\KwIn{Video features: $\bm{V}_{train}$ and $\bm{V}_{val}$; Individual Identification net: $\mathcal{I}(X; \theta)$; Temporal Reasoning net: $\mathcal{T}(X; \phi)$; Training epochs for $\mathcal{I}$ and $\mathcal{T}$: $N_{\mathcal{I}}$ and $N_{\mathcal{T}}$; Noise weights for $\mathcal{I}$ and $\mathcal{T}$: $\lambda_{\mathcal{I}}$ and $\lambda_{\mathcal{T}}$; Gaussian noise: $\bm \epsilon$}
\KwOut{The best model: $\mathcal{I}(\theta^{\prime})$ and $\mathcal{T}(\phi^{\prime})$}

\For{$j \in 1, 2, ..., N_{\mathcal{I}}$}{
$\theta, \phi = \arg \min_{\theta, \phi}L(\bm{V}_{train} + \lambda_{\mathcal{I}} \cdot \bm \epsilon; \theta, \phi)$

\text{Find the best accuracy}: $\text{eval}(\mathcal{I}(\bm{V}_{val}; \theta))$

}

\For{$i \in 1, 2, ..., N_{\mathcal{T}}$}{

$\phi = \arg \min_{\phi}L_{T}(\bm{V}_{train} + \lambda_{\mathcal{T}} \cdot \bm \epsilon; \theta, \phi)$

\text{Find the best edit}: $\text{eval}(\mathcal{T}(\mathcal{I}(\bm{V}_{val}; \theta); \phi))$

}
\end{algorithm}

\begin{figure}[t]
  \centering
  \includegraphics[width=\linewidth]{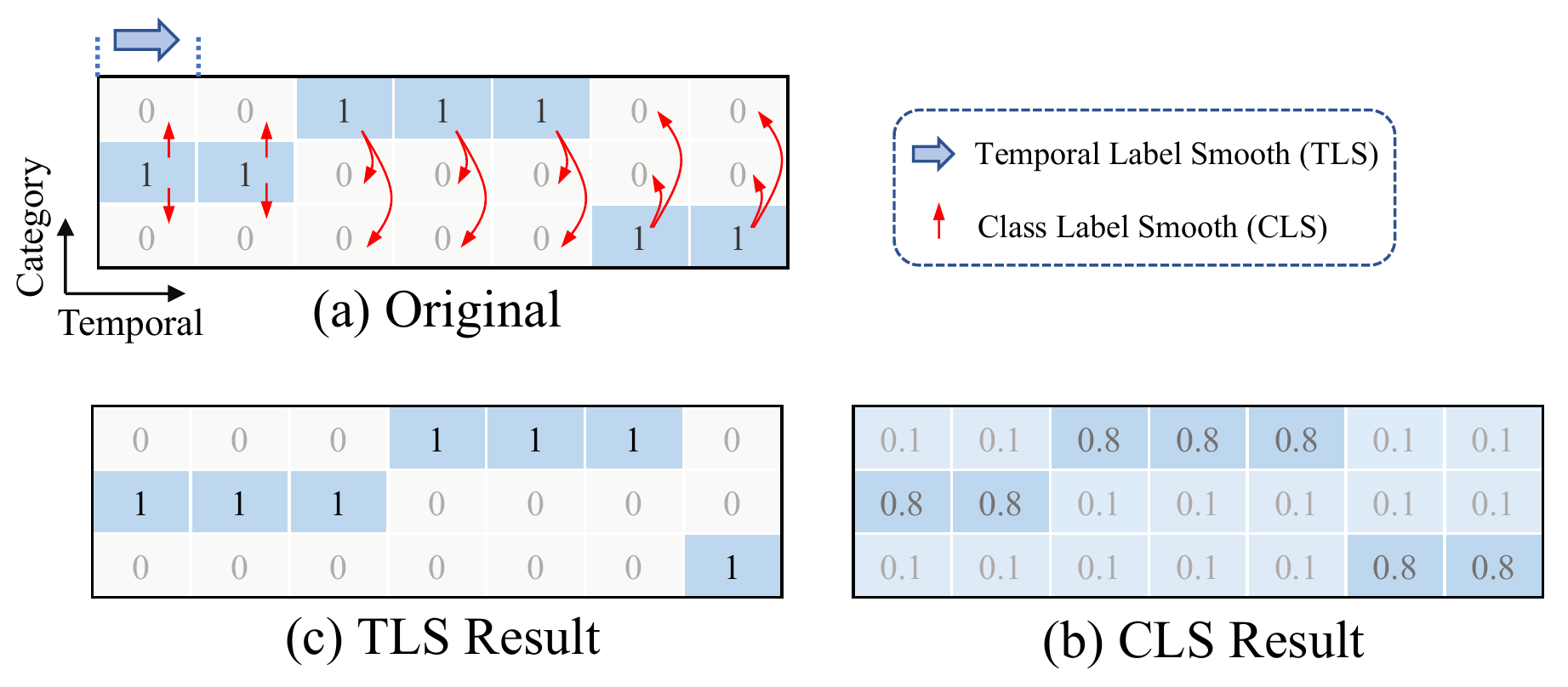}
  \caption{The original one-hot encoding (a) is commonly used in cross-entropy loss. TLS (b) is applied on the temporal channel. We can see that only the boundary targets will be changed to the adjacent labels. CLS (c) is applied on the categorical channel, which ensures the total of all the classes still to be one.}
  \label{fig:ls}
\end{figure}

\subsection{Temporal and categorical label smoothness}
\begin{figure}[t]
  \centering
  \includegraphics[width=0.485\textwidth]{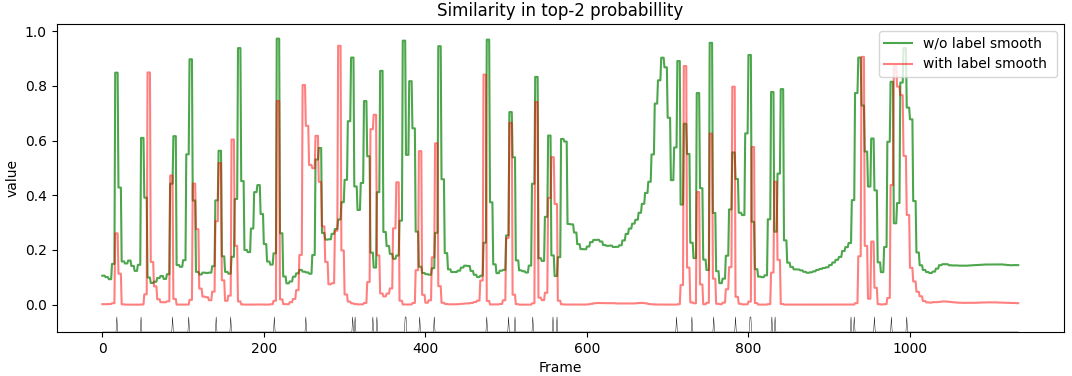}
  \caption{The similarity of top-2 probabilities among a video before and after using label smoothness. The short black lines at the bottom are the locations of action boundaries.  }
  \label{fig:top2}
\end{figure}

Because of the boundary ambiguity, the action classifier is more likely to produce some unreliable predictions nearby action boundaries. Interestingly, we find that the training losses near the action boundaries cannot be as low as the non-boundaries, shown in Figure~\ref{fig:top2}. We further visualize the similarity of the top 2 probabilities for each frame and observe prediction ambiguity near the boundaries. To remedy this problem, a potential solution is boundary-aware supervision~\cite{wang2020boundary}. In our proposed method, we simplify the solution by rolling the action transcript temporally to allow the mapping between inputs and labels nearby action boundaries \textbf{not} to be precisely aligned.

\noindent\textbf{Temporal Label Smoothness.} We propose a Temporal Label Smoothness (TLS) operation during training. It rolls the target labels temporally by a step size uniformly sampled from 1 to $\epsilon$, where $\epsilon$ is a hyper-parameter. Obviously, it does not affect the targets on the non-boundary segments.  We empirically find that such a simple strategy effectively tackles hard samples caused by certain labeling on continuous features. \textbf{More illustration can refer to Section~\ref{analysis}.}

\noindent\textbf{Categorical Label Smoothness.}
Besides, visually similar action classes also introduce hard examples to the classifier trained on the one-hot encoding labels. Thus, we adopt the commonly used categorical label smoothness in both individual identification and temporal reasoning model with hyper-parameter $\beta$ and $\beta'$, respectively, to relax the one-hot target encoding to a float encoding also with a total of one. 

We illustrate the temporal label smoothness and categorical label smoothness in Figure~\ref{fig:ls}.

\subsection{Objective Functions}
\noindent\textbf{Frame-wise classification Loss.} We use the cross-entropy loss for frame-wise prediction $\hat{p}_{i, c}$ with ground-truth class $c$ in individual identification:
\begin{equation}
\begin{aligned}
    L_I = -\sum_{i=1}^{T}\log(\hat{p}_{i, c})
\end{aligned}
\end{equation}
\textbf{Action Transcript Loss.} The cross-entropy loss is also used for action transcript prediction in temporal reasoning:
\begin{equation}
\begin{aligned}
    L_T = -\sum_{i=1}^{N_a}\log(\hat{a}_{i, c})
\end{aligned}
\end{equation}
where $N_a$ is the number of ground truth actions in each video, and $a_i$ is the action-wise prediction with ground-truth class $c$.

\noindent\textbf{Total Loss.} These two losses are combined equally:
\begin{equation}
\begin{aligned}
    L = L_I + L_T
\end{aligned}
\end{equation}

\setlength{\abovecaptionskip}{0.1cm}
\begin{table*}[t]
  \caption{Comparison with recent state-of-the-art methods.}
  \label{sota}
  \centering
  \small
   \resizebox{.99\linewidth}{!}{
  \begin{tabular} {@{\hspace{.5mm}}l@{\hspace{.5mm}} |c|c|ccccc|ccccc}
    \hline
     & & & \multicolumn{5}{c}{GTEA} & \multicolumn{5}{|c}{Breakfast} \\ \hline
    \multicolumn{1}{c}{Method} & \multicolumn{1}{|c}{Architecture} &\multicolumn{1}{|c}{Feature}&\multicolumn{3}{|c}{F1@\{10, 25, 50\}} & Edit & Acc.(\%)&\multicolumn{3}{c}{F1@\{10, 25, 50\}} & Edit & Acc.(\%)\\    \hline
    ED-TCN~\cite{lea2017temporal}& CNN&-& 72.2 & 69.3 & 56.0 &- &64.0 &  - & -&- &- &43.3   \\
  
    TDRN~\cite{lei2018temporal}& CNN&- & 79.2  & 74.4 & 62.7 & 74.1 & 70.1 &   -   & - & - & - & - \\
    SSA-GAN~\cite{gammulle2020fine}& CNN & GAN & 80.6 & 79.1& 74.2 &76.0 & 74.4 &  - & -&- &- &-   \\
    DA~\cite{chen2020action} & CNN & - &74.2 &68.6& 56.5& 73.6& 71.0  & 74.2 & 68.6 & 56.5 & 73.6 & 71.0 \\
    MS-TCN++~\cite{li2020ms}& CNN & Multi-stage&88.8  & 85.7 & 76.0 & 83.5 & 80.1 & 64.1 &58.6 & 45.9 & 65.6 & 67.6\\
    BCN~\cite{wang2020boundary} & CNN & Boundary & 88.5 &87.1& 77.3& 84.4& 79.8  & 68.7 &65.5& 55.0& 66.2& 70.4\\
    DTGRM~\cite{wang2021temporal}& GCN & - & 87.3 & 85.5 & 72.3 & 80.7 & 77.5 &  68.7& 61.9 & 46.6 & 68.9 &68.3   \\
    Gao \etal ~\cite{gao2021global2local}& CNN & NAS & 89.9 & 87.3 & 75.8 & 84.6 &78.5 & 74.9 & 69.0 & 55.2 & 73.3 & 70.7   \\
    ASRF~\cite{ishikawa2021alleviating} & CNN & Boundary& 89.4 & 87.8 & 79.8 & 83.7 & 77.3 & 74.3 & 68.9 & 56.1 & 72.4 & 67.6\\
    ASFormer~\cite{yi2021asformer} & Transformer & Multi-stage & 90.1 & 88.8 & 79.2 & 84.6 & 79.7 & \underline{76.0} &\underline{70.6} & \underline{57.4} & 75.0 & \underline{73.5}\\
    UVAST+Viterbi~\cite{uvast2022ECCV} &Transformer& Unification & \underline{92.7} & \underline{91.3} & \underline{81.0} & \textbf{92.1} & \underline{80.2} & 75.9 & 70.0 & 57.2 & \underline{76.5} & 66.0\\
    \hline
    DIR-AS &Transformer & Decoupling &  \textbf{94.3} & \textbf{93.4} & \textbf{84.5} & \underline{89.7} & \textbf{82.8} & \textbf{76.2} & \textbf{71.0} & \textbf{58.2} &\textbf{76.6} & \textbf{74.7} \\
    \hline
  \end{tabular}}
\end{table*}

\begin{table*}
  \begin{minipage}{.63\linewidth}
    \centering
        \resizebox{.97\linewidth}{!}{\begin{tabular}{c|ccccc|ccccc}
            \hline
            \multicolumn{1}{c}{} &
            \multicolumn{5}{|c|}{Module} & \multicolumn{5}{c}{GTEA (Split 1)} \\
            \hline 
            \multicolumn{1}{c|}{} & 
            {\makecell[c]{Conv.\\ Stem}} & {\makecell[c]{Local\\ branch}} & {\makecell[c]{Global\\ branch}} & {\makecell[c]{Label\\ Smooth.}} & {\makecell[c]{Temporal\\ Reasoning}} &
            \multicolumn{3}{c}{F1@\{10, 25, 50\}}  & \multicolumn{1}{c}{Edit} & \multicolumn{1}{c}{Acc.(\%)}\\
            \hline
            a &  & \ding{51} &  &  & & 50.00 & 44.49  & 34.86  & 41.51 & 71.20 \\
            b &  & \ding{51} &\ding{51}  &  & & 45.95 & 42.01 & 35.01  &  39.26  & 73.47  \\
            c & \ding{51} & \ding{51} & \ding{51} & & & 80.00 & 78.03 & 69.32  & 71.45  & 77.43 \\
            d & \ding{51} & \ding{51} & \ding{51} & \ding{51} & & 85.00  & 82.85 & 70.71 &   79.84 &  79.39 \\
            e & \ding{51} & \ding{51} & \ding{51} & \ding{51} &\ding{51} & \textbf{91.67} & \textbf{91.67} & \textbf{83.33} & \textbf{94.77} & \textbf{84.27} \\
            \hline
          \end{tabular}}
    \caption{Results of appending different modules in series. }\label{tab:module}
  \end{minipage}%
  \hfill
  \begin{minipage}{.35\linewidth}
    \centering
        \resizebox{\linewidth}{!}{ \begin{tabular}{c| c}
            \hline
             {\makecell[c]{Transformer-based\\ Models}} &{\makecell[c]{Running Time\\ (s/video)}}\\
            \hline
            ASFormer~\cite{yi2021asformer} & 0.1961  \\
            UVAST (+ alignment decoder)~\cite{uvast2022ECCV}  & 0.1400 \\
            DIR-AS (w/o Viterbi) & \textbf{0.0439}\\
            \hline
          \end{tabular}}
    \caption{Running time of different methods on GTEA.}\label{tab:runtime}
  \end{minipage}
\end{table*}

\subsection{Input Augmentation}
As the inputs of the network are visual features rather than raw images, we augment the inputs by adding Gaussian noise to alleviate over-fitting. It facilitates the training of the temporal reasoning network as it slightly alters the coarse transcripts yielded from the individual identification network, as illustrated in Figure~\ref{fig:overall}. Note that we cannot directly add noise to the coarse transcripts, \ie, the input of temporal reasoning module, because it may result in a collapse of transcripts so that the network cannot normally converge.

\subsection{Decoupled Training}
The proposed individual identification network denoted by $\mathcal{I}$ works for frame-wise classification, while the temporal reasoning network denoted by $\mathcal{T}$ is good at segment-level prediction without action duration, \ie, transcripts. However, end-to-end training on these two tasks with different goals always causes a sub-optimal resolution for all the evaluation metrics focusing on different granularity, \eg, frame-wise accuracy, segment-wise F1 score, and segment-level Edit score. Thus, we propose a decoupled training strategy for our proposed DIR-AS in Algorithm~\ref{alg:alg1}. First, we perform end-to-end training by combining both networks.  Here $\mathcal{T}$ acts as an auxiliary task for training $\mathcal{I}$ as we intend to find the best accuracy. Interestingly, we find that it significantly facilitates the training of $\mathcal{I}$, compared to the isolated training. Afterward, we freeze $\mathcal{I}$ and fine-tune the $\mathcal{T}$ to find the best Edit score. 
Note that here we use a larger noise weight, \ie, $\lambda_{\mathcal{T}} > \lambda_{\mathcal{I}}$ as we need to increase the input diversity of the conditioned label sequences derived the frozen $\mathcal{I}$. 
After decoupled training, we can use parameter-free action alignment to fuse these two network outputs or simply take the frame-wise classification of $\mathcal{I}$ as final predictions.

\section{Experiment}
\subsection{Dataset and Evaluation Metric}
We conduct all the experiments on two popularly tested datasets: GTEA~\cite{fathi2011learning},  and Breakfast~\cite{kuehne2014language}, which consist of 28 and 1712 videos with 11 and 48 action categories, respectively. Following the common settings, we perform 4-fold cross-validation. All ablation studies are conducted on the split one of GTEA as in previous work~\cite{ishikawa2021alleviating,li2020ms, wang2020boundary,yi2021asformer}. Compared to GTEA, Breakfast holds long-term videos, which are easier to yield over-segmentation for all the methods. For example, we observe inferior Edit scores and F1 scores when only training the individual identification. The same observation is reported in previous works~\cite{wang2020boundary, farha2019ms, yi2021asformer}.

Following the previous works~\cite{farha2019ms,li2020ms,wang2020boundary,ishikawa2021alleviating}, we use the accuracy over all frames and report the F1 score with different IoU thresholds, 0.10, 0.25, and 0.50, and segmental Edit score. Note that accuracy accounts for individual classification while the F1 score and Edit score account for temporal reasoning due to the action instantiation.

\subsection{Experiment Details} We use pre-trained I3D~\cite{carreira2017quo} to project each video to visual features with dimension $D=2048$, to obtain a condensed video representation capturing the appearance and motion patterns of video clips.

For convolutional downsampling at the beginning, window size 7 with 4$\times$ downsampling is applied. For individual identification, 
we employ three local-global layers. In each layer, the embedding dimension of sliding window attention with TPD is 512, and the number of heads is 9. For the local branch, the dilation rates are 1, 2, and 4, meaning that each dilation includes three heads. According to the average length of videos and segments of different datasets, the window size is set to 7 and 51 for GTEA and Breakfast, respectively. For the global branch, video features are downsampled by three average pooling in temporal dimension with $15\times$/$30\times$/$45\times$, which are almost equal to 1/2/3s.
For temporal reasoning, eight layers of Multi-Heads Self-Attention are applied. The input query tokens of this module, \ie coarse transcripts from the individual identification, are equipped with the `End' token and padded by the max length of 41 and 100 for GTEA and Breakfast, respectively.

During training the temporal reasoning network, we use ground-truth transcripts at the first 50 epochs to learn an identity mapping for convergence. Then, we replace it with predicted transcripts from individual identification. Moreover, the Gaussian noise $\bm \epsilon\sim N(0,1)$ are added to input features with weight $\beta=0.5$ and $\beta'=1$ before and after decoupled training, respectively. The losses with an Adam~\cite{kingma2014adam} optimizer are combined with different weights, \eg, 0.9 for the individual identification loss and 0.1 for the temporal reasoning loss. The learning rate is 0.0001, and the batch size is 1 for all the datasets.

\subsection{Comparison with State-of-the-art Methods}
We compare our proposed method, \ie, DIR-AS, with all the recent state-of-the-art methods in Table~\ref{sota}. We report much promising performance on GTEA and Breakfast. Specifically, DIR-AS achieves the best accuracy and F1 score on GTEA, which significantly outperforms the latest work UVAST~\cite{behrmann2022unified} \textbf{2.6\%} on the accuracy and \textbf{3.5\%} on the F1@50 score, respectively. For Breakfast, DIR-AS has a superior accuracy (74.7\%), compared to ASFormer (73.5\%). 
In addition, DIR-AS outperforms those multi-stage networks with frame-wise supervision, \eg, MS-TCN++~\cite{li2020ms} and ASRF~\cite{ishikawa2021alleviating} on the two datasets, which shows the effectiveness of decoupling the individual identification and temporal reasoning for action segmentation.

\subsection{Ablation Studies}

In this study, we have explored the impact of various modules on the GTEA dataset in Table~\ref{tab:module}. Our results demonstrate a gradual improvement in performance as more modules are added in series, indicating their complementarity to each other, ultimately leading to state-of-the-art accuracy when all modules are appended.

A comparison between (a) and (b) indicates an improvement in accuracy but a slight degradation in the F1 score and Edit score. It means although the global branch enlarges receptive fields to improve model accuracy, it weakens locality, a crucial factor for the prediction smoothness. Hence, the addition of the convolutional stem, as in (c), increases the locality, resulting in significant improvements in the F1 score ($\sim$ 40\%) and Edit score ($\sim$ 30\%), and a slight improvement(1.92\%) in accuracy.

Remarkably, the introduction of the label smoothness strategy improves all the metrics. We will provide more analysis in Section~\ref{analysis}. 
The addition of the transcript prediction module, as seen in (e), dramatically improves the F1@50 score by ~13\%, Edit score by ~15\%, and accuracy by ~5\%, respectively, which emphasizes the importance of the temporal reasoning module.

\subsection{Running Time}
\label{runtime}
Our study evaluates the running time of three transformer-based models for action segmentation: ASFormer, UVAST, and our proposed DIR-AS. The running time of each method is tested on GTEA split 1. We focus on end-to-end training models without post-processing. Thus, we select UVAST with an alignment decoder and our DIR-AS without Viterbi.
As shown in Table~\ref{tab:runtime}, ASFormer and UVAST have similar running times due to their adoption of similar transformer structures. Our proposed method takes less time and achieves the best accuracy, which demonstrates the efficiency of DIR-AS. To conduct running time testing, we utilized the official code implementations of methods~\cite{yi2021asformer,uvast2022ECCV}, and the experiments were executed on an NVIDIA RTX A5000 GPU.


\subsection{More Analysis}
\label{analysis}

\noindent\textbf{How to design the temporal reasoning network reasonable?} 
We conduct more studies on the temporal reasoning module, which is achieved based on a Query-Key-Value attention manner. Specifically, the key tokens are from the deep or shallow layers in the individual identification module, \ie,  the features before classification denoted as ``Deep'', and the features after convolutional stem denoted as ``Shallow'', as shown in Table~\ref{tab:pos}. The query tokens, on the other hand, can be either conditioned from individual identification, \ie, segment-wise probability (Prob.), predicted transcript (Trans.), or unconditioned learned tokens (Learned).

For the selection of the query tokens, the learned tokens are not the best choice with both ``Deep'' or ``Shallow'' keys. We assume the unconditioned learned tokens provide poor prior for temporal reasoning, especially on the small dataset that is unable to provide the inductive bias. However, the segment-wise probability and predicted transcript help temporal reasoning as they are conditional on each input sample. Remarkably, we observe a significant improvement on accuracy, \ie, 4.7\%, when adopting predicted transcripts as queries to attend ``Deep'' and ``Shallow'' features, respectively. It suggests that less network sharing between identification individual and temporal reasoning will be helpful for each other, which supports our decoupling motivation.

\begin{figure*}[htb]
\centering
{\includegraphics[width=0.485\linewidth]{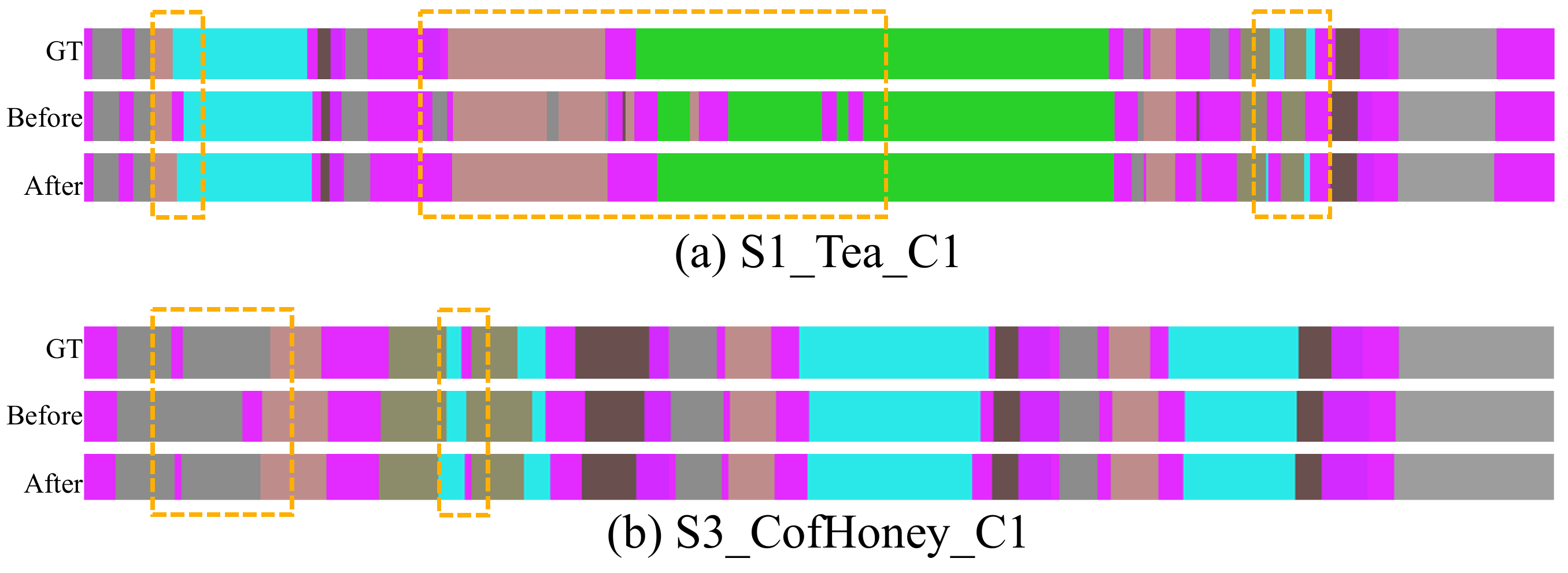}}
\hfill
{\includegraphics[width=0.485\linewidth]{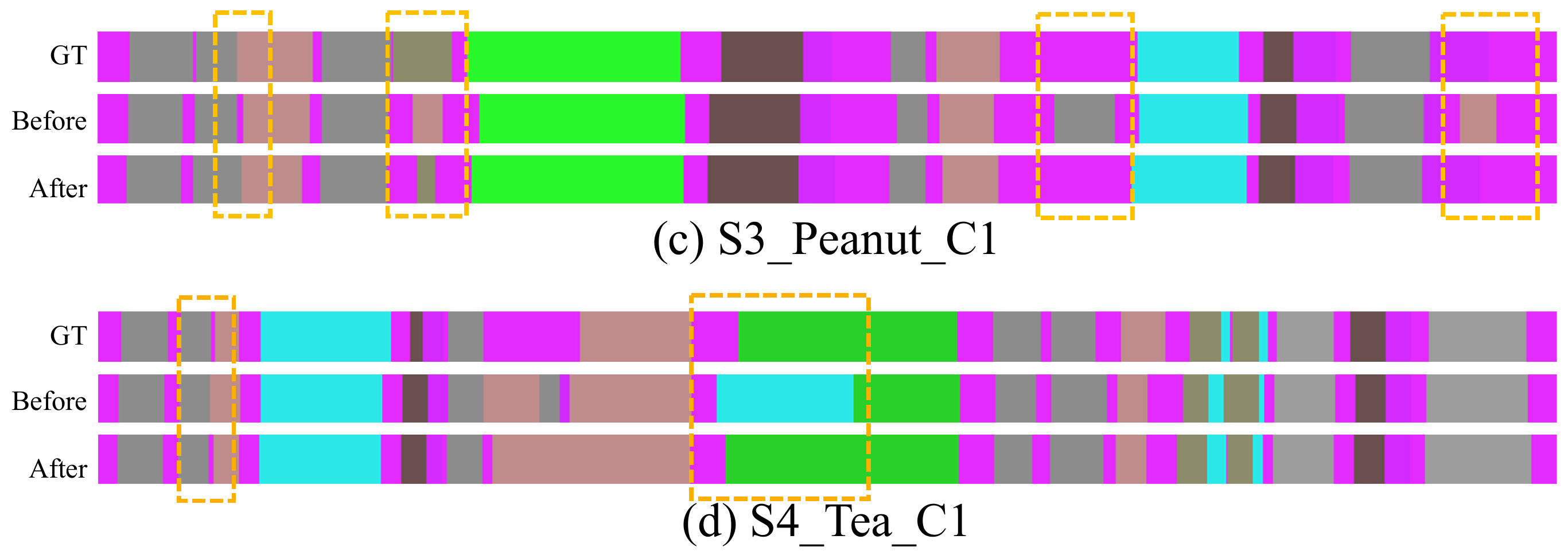}}
\caption{Comparison of GTEA results with and without Viterbi fusion. The figure displays three rows for each sample: (1) ground truth, (2) DIR-AS before Viterbi fusion, and (3) DIR-AS after Viterbi fusion with action alignment. The yellow boxes highlight cases where the initial individual identification results were incorrect, but were corrected through the fusion process, which involved recalling, removing, and updating action segments.  }
\label{fig:vis}
\vspace{-2mm}
\end{figure*}

\begin{figure}[t]
\centering
\subfloat[Accuracy with or without training noise.]{\includegraphics[width=0.48\linewidth, height=.39\linewidth]{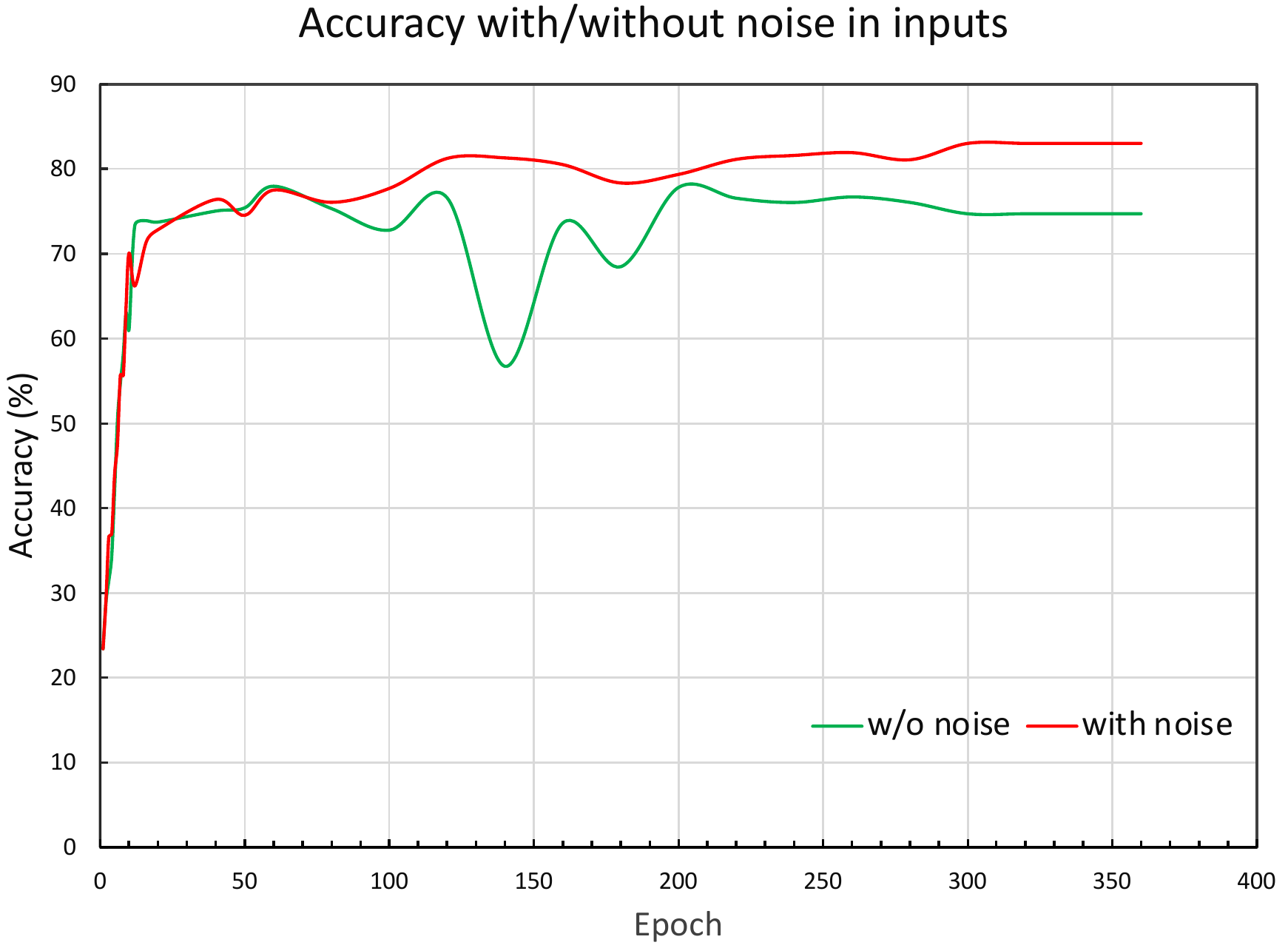}}
\hfill
\subfloat[Edit with or without training noise.]{\includegraphics[width=0.48\linewidth, height=.39\linewidth]{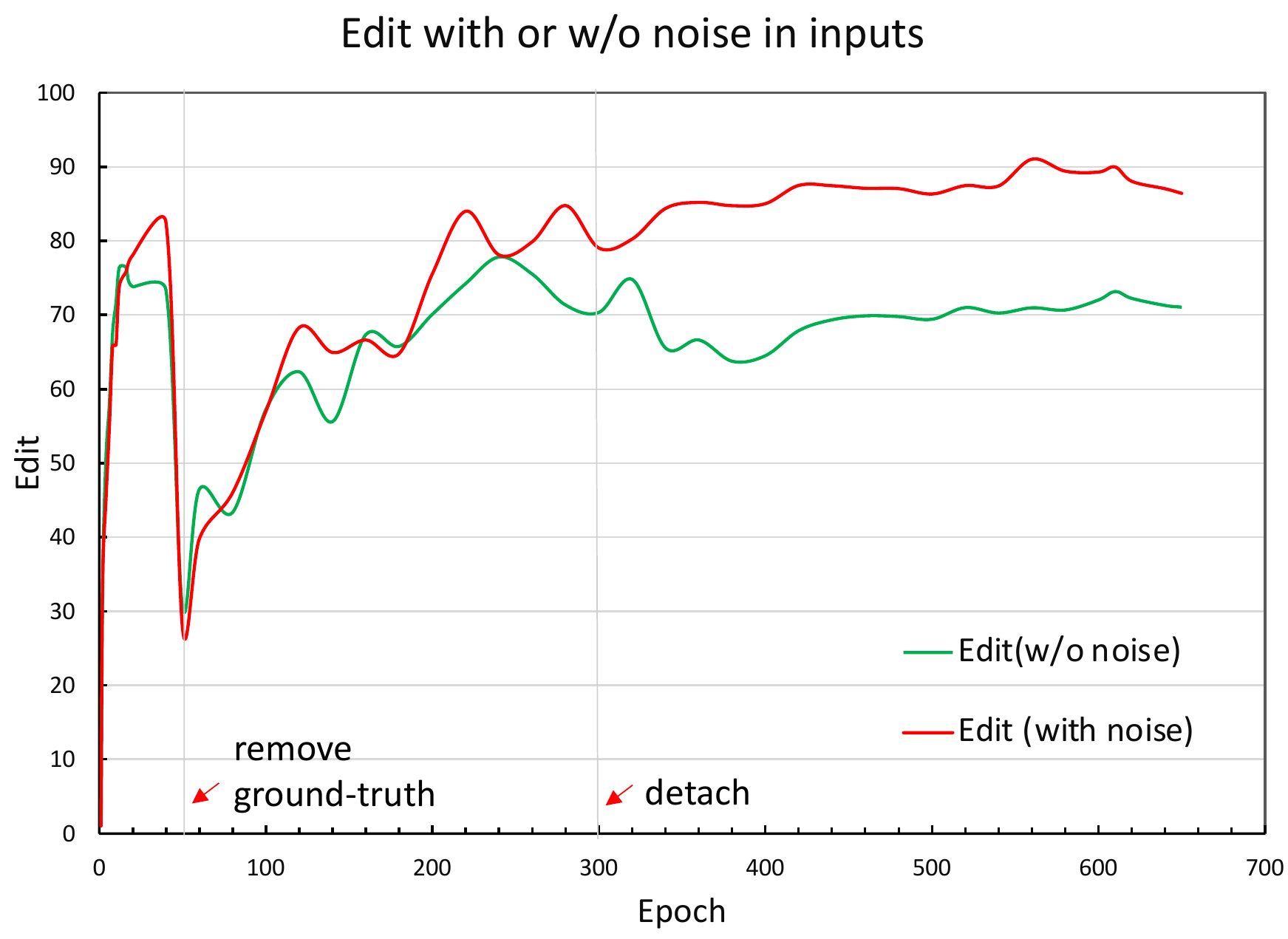}}
\caption{The effect of adding Gaussian noise for training in inputs.}
\label{fig:noise}
\vspace{-4mm}
\end{figure}

\noindent\textbf{How does the noise help the training?} 
Remind that we augment input features by adding Gaussian noise in Algorithm~\ref{alg:alg1}. To investigate the efficacy of this technique, we present the metric curves on the validation set in conjunction with the training process in Figure~\ref{fig:noise}. By adding noise to input features, the accuracy of the validation exhibits a gradual increase, while it becomes unstable without feature augmentation. Similar observations are noted for the Edit score. This can be explained by the fact that the model is forced to be more robust and adaptive to variations in the input data, which helps to prevent overfitting and improve the generalization performance on validation.

Moreover, the introduction of Gaussian noise endows the temporal reasoning module with the capability to attain a superior Edit score, as we observe a significant improvement in the Edit score after decoupling training. This is because the predicted transcripts from the frozen individual identification will also be augmented by adding noise to input data, avoiding sub-optimal optimization for temporal reasoning to some extent.

\begin{table}[t]
\centering
  \caption{Results of using different query and key tokens in MHCA for temporal reasoning on GTEA. Keys are derived from features of shallow or deep layers, while queries include learned tokens randomly initialized, predicted probability, and transcripts.}
  \label{tab:pos}
  \small
  \resizebox{.99\linewidth}{!}{
  \begin{tabular}{c|c|c|ccccc}
    \hline
    & Key & Query  &
    \multicolumn{3}{c}{F1@\{10, 25, 50\}}  &  Edit& Acc.(\%)\\
    \hline
    a & {\multirow{3}{*}{Deep}} & Learned  & 86.45 & 84.25 & 75.46 & 78.60& 79.76\\
    b &  & Prob.  & 87.36 & 85.92 & 72.20 & 82.19 & 80.85\\
    c &  & Trans.  & 87.68 & 85.51 & 75.36  & 80.96& 79.53\\
    \hline
    d &{\multirow{3}{*}{Shallow}} & Learned   & 83.94 & 79.56 & 67.15 & 78.67 & 78.20\\ 
    e & & Prob.  & 87.27 & 85.82  &  74.18 & 76.04 & 80.35\\
    f & & Trans.  &\textbf{91.67} & \textbf{91.67} & \textbf{83.33} & \textbf{94.77} & \textbf{84.27}\\
    \hline
  \end{tabular}}
  \vspace{-4mm}
\end{table}

\noindent\textbf{How does label smoothness help to reduce boundary ambiguity?} The presence of sudden switches in the annotations at the boundaries, despite the continuous nature of video features, introduces ambiguity into the training process. In order to elucidate the efficacy of our proposed solution, namely label smoothness, we introduce a metric to quantify the level of ambiguity, based on the similarity between the top two predicted probabilities. It is important to note that higher similarity values correspond to a higher degree of the inseparability of the classifier. As depicted in Figure~\ref{fig:ls}, due to the absence of label smoothness, high inseparability frequently occurs near action boundaries, as indicated by the green line being consistently above the red line. Conversely, when the label smoothness is applied temporally and categorically, the similarities near the boundaries are significantly reduced. This can be attributed to the fact that label smoothness allows the boundary features to be attached to two action labels, leading to an average prediction over these two actions. Furthermore, label smoothness does not introduce any additional parameters, which is computationally efficient.

\subsection{Visualization.} In Figure~\ref{fig:vis}, we also visualize the predicted and ground-truth segmentations of our proposed DIR on GTEA. Thanks to the decoupling strategy, DIR shows continuous predictions. Without the temporal reasoning branch, our proposed method yields over-segmentation as expected, which well supports our motivation in this paper. After using the action alignment to fuse the two network predictions, the old over-segmentation problem can be alleviated. Additionally, in the second box of Figure~\ref{fig:vis} (b), a tiny action is recalled after fusion predictions, which demonstrates the fine-grained recognition capacity of the proposed temporal reasoning network. However, we also observe some failure cases where a single segment is split into multiple sub-segments, which means a new over-segmentation issue. Overall, we observe a complementary between the fusion results and the isolated results from the individual identification network. More visualization can be found in the supplementary.  

\section{Conclusion}
In this paper, we develop a two-branch Transformer for action segmentation with two decoupled goals, \ie, individual identification and temporal reasoning. For the first one, we design a dedicated efficient local-global attention equipped with multi-scale dilated window attention and multi-scale global attention for large receptive fields. For the second one, we introduce a conditional transcript prediction network that only produces action-wise predictions. Parameter-free action alignment is used to fuse these two branch predictions. Our method presents superior performance on two challenging datasets. We expect this work to bring some novel insights for action segmentation and other fine-grained video tasks.

\begin{center}
{\Large \bf Appendix}
\end{center}

\renewcommand{\thesection}{\Alph{section}} 
\setcounter{section}{0}
\section{Dataset Statistic}
Table~\ref{tab:stat} provides a comprehensive summary of the datasets employed in our experiments, namely GTEA~\cite{fathi2011learning} and Breakfast~\cite{kuehne2014language}. It is worth noting that the Breakfast dataset contains a greater number of classes and videos than GTEA. Nonetheless, GTEA comprises a higher number of action segments in each video, while Breakfast exhibits longer and less segmented videos. These differences underscore the significance of meticulous consideration of dataset properties when performing experiments and selecting hyperparameters for the proposed model.

\begin{table}[h]
\centering
  \caption{Data statistic of datasets on average video length, segments per video and segments length. }
  \label{tab:stat}
  \small
     \resizebox{.95\linewidth}{!}{
  \begin{tabular}{c|c|c|c|c|c }
    \hline
    Dataset & \makecell{\#action\\class} & \#videos & \makecell[c]{video \\length(min)}&  \makecell[c]{ \#segments \\per video} & \makecell[c]{ segments\\ length(s) } \\
    \hline
     GTEA~\cite{fathi2011learning} & 11 & 28 & 1.24 & 31 & 2.21 \\
    Breakfast~\cite{kuehne2014language}& 48 & 1712 & 2.3 & 6.6 & 15.1\\
    \hline
  \end{tabular}}
\end{table}

\section{Details of Hyperparameters}

\begin{table}[tb]
\centering
  \caption{Hyper-parameters of DIR-AS for datasets of GTEA and Breakfast.}
  \label{tab:params}
  \small
  \begin{tabular}{c|c|c|c }
    \hline\hline
     & Hyper-parameter & GTEA &  Breakfast \\\hline\hline
   &  max. length & 50 & 100\\\cline{2-4}
   
   &  \# local\_global &\multicolumn{2}{c}{3}\\ \hline \hline
   \multirow{3}{*}{\makecell{Convolutional\\Stem}}& embed dim. &\multicolumn{2}{c}{512}\\\cline{2-4}
   & patch kernel size & \multicolumn{2}{c}{7}\\\cline{2-4}
   & patch stride & \multicolumn{2}{c}{4}\\\cline{2-4}
   & \# conv. layer& \multicolumn{2}{c}{3}\\\cline{2-4}
   & conv. kernal size & \multicolumn{2}{c}{3}\\\cline{2-4}
   & conv. dilation & \multicolumn{2}{c}{[1, 2, 4]}\\\hline\hline
   \multirow{5}{*}{\makecell{Local\\Branch}} & \#head & \multicolumn{2}{c}{9}\\\cline{2-4}
   &head dim. & \multicolumn{2}{c}{128}\\\cline{2-4}
   &hidden dim. & \multicolumn{2}{c}{1024}\\\cline{2-4}
   &head dilation & \multicolumn{2}{c}{[1, 2, 4]}\\\cline{2-4}
   &window size & \ 7 & 51\\\hline\hline
   \multirow{4}{*}{\makecell{Global\\Branch}}    & \#heads & \multicolumn{2}{c}{9}\\\cline{2-4}
    & head dim. & \multicolumn{2}{c}{128}\\\cline{2-4}
    & hidden dim. & \multicolumn{2}{c}{1024}\\ \cline{2-4}
    & window size & \multicolumn{2}{c}{[15, 45, 90]}\\ \hline\hline
    \multirow{4}{*}{\makecell{Temporal\\ Reasoning\\Module}} & input dim. & \multicolumn{2}{c}{512}\\ \cline{2-4}
    & head dim. & \multicolumn{2}{c}{128}\\ \cline{2-4}
    & hidden dim. &\multicolumn{2}{c}{512}\\ \cline{2-4}
    & \# self-attention layer &\multicolumn{2}{c}{8}\\ \hline
    \hline
    \multirow{3}{*}{\makecell{Label\\Smoothness}}&  \makecell{TLS (Individual \\identification )} & 4 & 10 \\ \cline{2-4}
   &  \makecell{CLS (Individual\\ identification) }& 0.1 & 0.2 \\ \cline{2-4}
   & \makecell{CLS (Temporal \\ressoning)}  & 0.4 & 0.45\\ \hline\hline
  \end{tabular}
\end{table}

Table~\ref{tab:params} provides an overview of the hyperparameters utilized in the training of our model. The convolutional stem involves the use of convolutional layers to obtain feature embeddings with temporal locality. In particular, we first apply a convolutional layer with a kernel size of 7 and a stride of 4, resulting in a $4\times$ downsampling. Subsequently, three convolutional layers with dilation rates of 1/2/4 are applied to increase the receptive field while preserving locality. The individual identification module consists of both local and global branches, with the local branch employing a window size of 7 for the GTEA dataset, which has shorter segments, and a window size of 51 for the Breakfast dataset to retain as much segment information as possible. In the temporal reasoning module, we aggregate video frames to obtain global information via cross multi-head attentions, with all hyperparameters being identical for both datasets, except for the maximum length. To ensure a fair comparison with previous methods, we employ a sampling rate of 1 for all datasets. Additionally, class label smoothness(CLS) is applied to both the individual identification and temporal reasoning modules Furthermore, to mitigate the issue of boundary ambiguity, temporal label smoothness (TLS) is solely implemented in the individual identification module, and the Breakfast adopt larger value 10 due to its longer segments. This incorporation of CLS and TLS in our model is intended to regularize the training process and enhance the model's generalization ability by imposing constraints on the predicted class labels. It is important to note that these hyperparameters were chosen based on extensive experimentation and fine-tuning to maximize the model's performance on the given datasets.


\section{More visualization} 
\subsection{More Results on Breakfast}
\label{sec: bre}
\begin{figure}[h]
  \centering
  \includegraphics[width=0.48\textwidth]{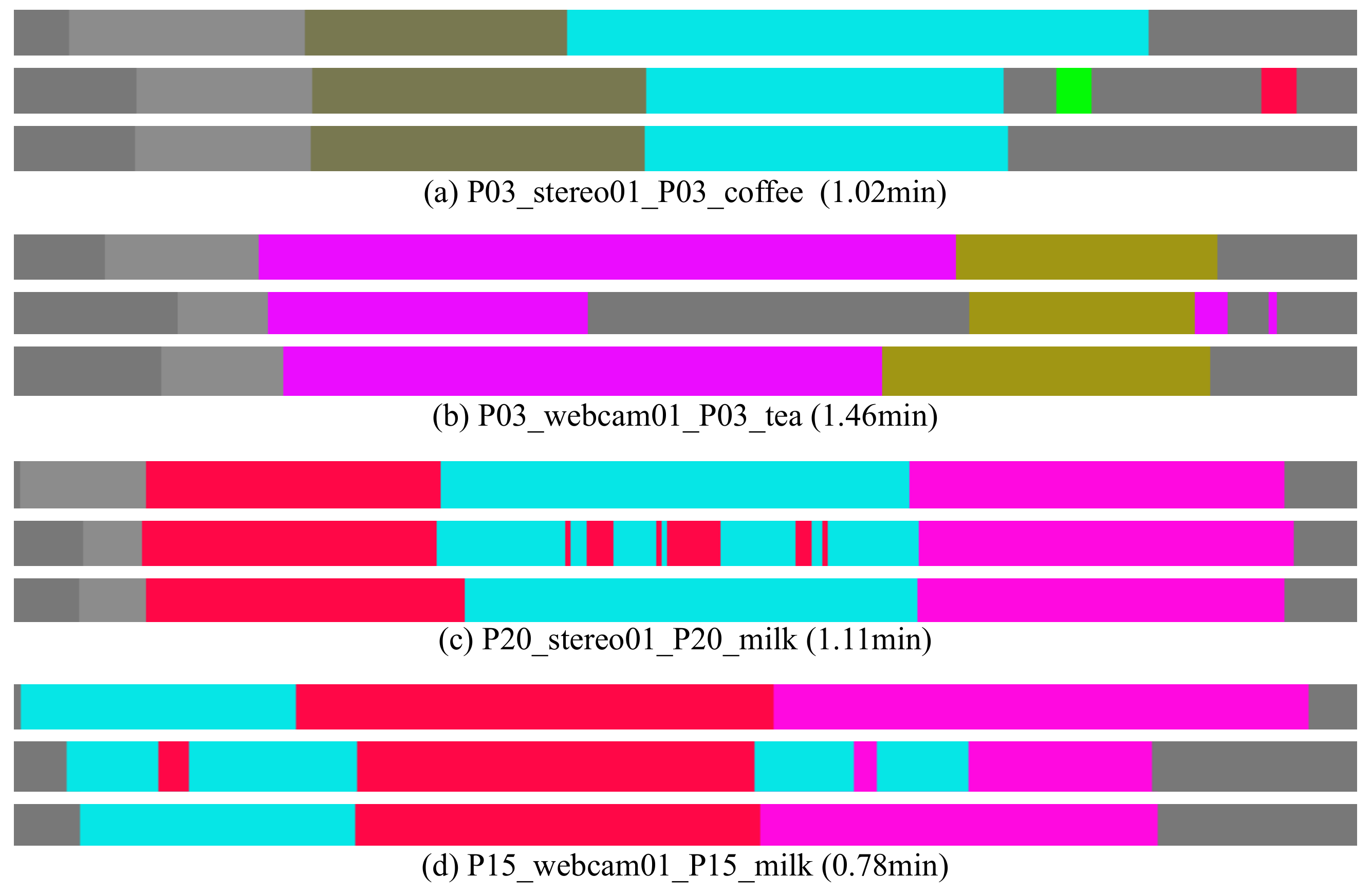}
  \caption{Visualization on Breakfast. From top to bottom, the visualizations are ground truth, DIR-AS before Viterbi fusion, and DIR-AS after Viterbi fusion. The time of videos are showed after. }
  \label{fig:bre_vis}
\end{figure}

We show the predictions in Figure~\ref{fig:bre_vis} before and after Viterbi fusion in the Breakfast dataset. Generally, the predictions before Viterbi fusion are close to ground truth except for the boundaries. However, in some longer videos with longer action segments, the over-segmentation problem still exists. 
By incorporating the predicted transcripts, which perform temporal reasoning by attending to the complete video, Viterbi fusion generates smoother results. 

\subsection{Failure Case}
\begin{figure}[h]
  \centering
  \includegraphics[width=0.48\textwidth]{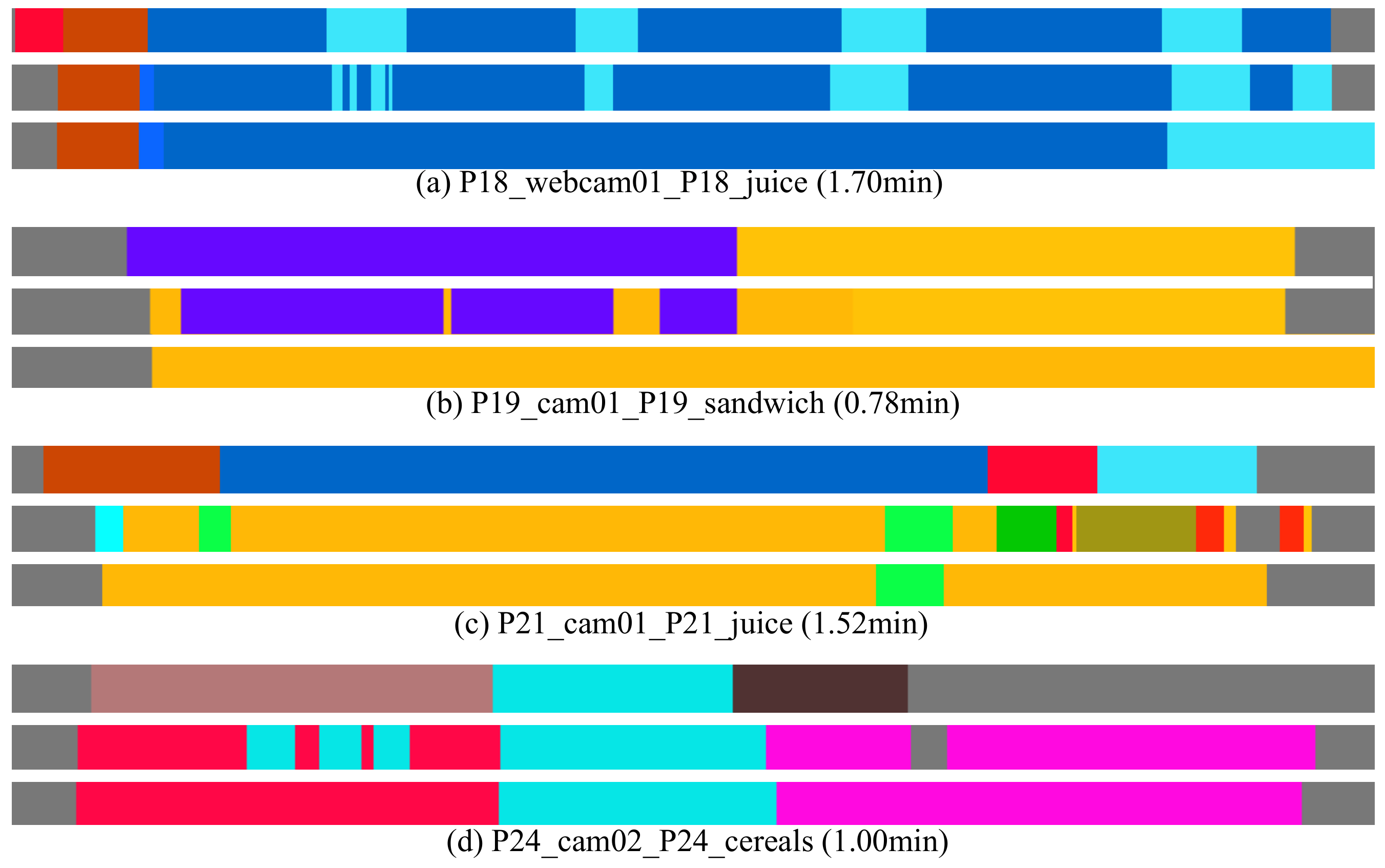}
  \caption{Failure cases. From top to bottom, the visualizations are ground truth, DIR-AS before Viterbi fusion, and DIR-AS after Viterbi fusion. }
  \label{fig:bre_fal_vis}
\end{figure}


Figure~\ref{fig:bre_fal_vis} displays several failure cases that require analysis. The present study focuses on evaluating the effect of alignment over DIR-AS. We identify two potential reasons for the deterioration of the predictions after alignment. Firstly, the transcript predictions may overlook some short segments, leading to under-segmentation. Secondly, compared to dense frame-wise prediction, sparse set prediction may suffer from insufficient training data. Specifically, in Figure~\ref{fig:bre_fal_vis}(a)(b), some short segments are eliminated by the alignment process, causing a loss of the originally existing actions. Conversely, in Figure~\ref{fig:bre_fal_vis}(c)(d), the alignment process can alleviate over-segmentation; however, the transcript predictions are similar to the frame-wise predictions, particularly in longer videos. These observations highlight the need for careful evaluation of alignment methods to ensure optimal results in action segmentation.

{\small
\bibliographystyle{ieee_fullname}
\bibliography{egbib}
}

\newpage

\end{document}